\documentclass[journal]{IEEEtran}
\usepackage{ifpdf}

\ifCLASSOPTIONcompsoc
  \usepackage[nocompress]{cite}
\else
  \usepackage{cite}
\fi
\ifCLASSINFOpdf
  \usepackage[pdftex]{graphicx}
  \graphicspath{{../pdf/}{../jpeg/}}
  \DeclareGraphicsExtensions{.pdf,.jpeg,.png}
\else
\fi
\usepackage{amsmath}
\usepackage{algorithmic}

\usepackage[ruled,norelsize]{algorithm2e}

\makeatletter
\newcommand{\removelatexerror}{\let\@latex@error\@gobble}
\makeatother

\usepackage{array}

  \usepackage[caption=false,font=footnotesize,labelfont=sf,textfont=sf]{subfig}
\usepackage{booktabs}
\usepackage{multirow}
\usepackage{xcolor}
\usepackage{color}
\usepackage{url}
\usepackage[normalem]{ulem} 
\usepackage{colortbl}

\newcommand{\etal}{\textit{et al}.}
\newcommand{\ie}{\textit{i}.\textit{e}.,}

\newcommand{\add}[1]{{#1}}
\newcommand{\del}[1]{}
\definecolor{myBlue}{rgb}{0,0,0}

\begin{document}
%
\title{Robust Iris Presentation Attack Detection Fusing 2D and 3D Information}
%
%
%
%
%
%

\author{Zhaoyuan Fang,~\IEEEmembership{Student Member,~IEEE,}
        Adam Czajka,~\IEEEmembership{Senior Member,~IEEE,}
        and~Kevin W. Bowyer,~\IEEEmembership{Fellow,~IEEE}
\IEEEcompsocitemizethanks{E-mails: \{zfang,aczajka,kwb\}@nd.edu
\IEEEcompsocthanksitem}
\thanks{Manuscript accepted on August 3, 2020}}

%
%

\markboth{Accepted for publication in the IEEE TRANSACTIONS ON INFORMATION FORENSICS AND SECURITY}%
{Fang \MakeLowercase{\textit{et al.}}: Robust Iris Presentation Attack Detection Fusing 2D and 3D Information}

%



\IEEEtitleabstractindextext{%
\begin{abstract}

 Diversity and unpredictability of artifacts potentially presented to an iris sensor calls for presentation attack detection methods that are agnostic to specificity of presentation attack instruments. This paper proposes a method that combines two-dimensional and three-dimensional properties of the observed iris to address the problem of spoof detection in case when some properties of artifacts are unknown. The 2D (textural) iris features are extracted by a state-of-the-art method employing Binary Statistical Image Features (BSIF) and an ensemble of classifiers is used to deliver 2D modality-related decision. The 3D (shape) iris features are reconstructed by a photometric stereo method from only two images captured under near-infrared illumination placed at two different angles, as in many current commercial iris recognition sensors. The map of normal vectors is used to assess the convexity of the observed iris surface. The combination of these two approaches has been applied to detect whether a subject is wearing a textured contact lens to disguise their identity. Extensive experiments with \textit{NDCLD'15} dataset, and a newly collected \textit{NDIris3D} dataset show that the proposed method is highly robust under various open-set testing scenarios, and that it outperforms all available open-source iris PAD methods tested in identical scenarios. The source code and the newly prepared benchmark are made available along with this paper.
\end{abstract}

}

\maketitle


\IEEEdisplaynontitleabstractindextext


\ifCLASSOPTIONpeerreview
\begin{center} \bfseries EDICS Categories: BIO-MODA-IRI, BIO-MODA-PAD \end{center}
\fi

\IEEEpeerreviewmaketitle

\section{Introduction}\label{sec:introduction}

%
%
%
%

\IEEEPARstart
%
{P}{resentation} attacks are the presentations to biometric sensors aiming at driving biometric systems into making incorrect decisions about one's identity. They could be used by attackers to fool the authentication process and obtain unauthorized access to information, either by impersonating a registered user or by concealing the identity of attackers. In iris recognition, possible forms of presentation attacks include using textured contact lenses, paper iris printouts, prosthetic eyes, or even cadaver eyes, since post-mortem iris recognition has been recently demonstrated to be possible \cite{Czajka_2018_PADsurvey}. In particular, several works~\cite{Baker2010DegradationOI, Kohli_2013_cosmetic, Yadav_2014_contactlens} have shown that wearing textured contact lens significantly degrades the performance of iris recognition systems. In addition, results from the LivDet-Iris 2017 competition~\cite{Yambay2017LivDet2017} show that state-of-the-art methods still offer limited detection accuracy, especially in open-set scenarios, where testing samples may vary from those seen in training. Depending on the database, the winning algorithm of LivDet-Iris 2017 still failed to detect 11\% to 38\% of presentation attacks. Therefore, iris Presentation Attack Detection (PAD) methods that generalize well 
in open-set scenarios
are crucial for iris recognition systems to be deployable in real-world applications. 

This paper builds upon our previous work~\cite{Czajka_2019_3DPAD}, which we refer to as OSPAD-3D. By fusing the OSPAD-3D with~\cite{McGrath_2018_OSPAD}, referred to as OSPAD-2D, we propose a novel open source iris PAD method, OSPAD-fusion, which unlike any previous work known to us exploits both 2D and 3D information of the iris from only two images to offer good generalization of detecting unknown textured contact lenses. Our method makes the prediction based on both 2D textural features from the lens patterns and 3D shape features from the normal map estimation. OSPAD-2D extracts Binary Statistical Image Features (BSIF)~\cite{Kannala_2012_BSIF} from iris images and uses an ensemble of classifiers to make predictions, while OSPAD-3D leverages the difference in shadows cast by contact lens on the iris surface, and uses normal maps estimated from photometric stereo to perform the classification. \add{The main motivation of this work is to offer an iris presentation attack method that is more agnostic to particular properties of textured contact lenses. As demonstrated later in the paper, such features change across different brands but also change in time for the same contact lens brand.}

In addition, as part of this effort we collected and offer a new dataset \textit{NDIris3D}, which contains images of both authentic irises and irises wearing contact lens taken by LG IrisAccess 4000 sensor (LG4000) with two different illuminants, and IrisGuard AD 100 sensor (AD100) with three different illuminants, under different illumination setups offered by these sensors. 
This new dataset allows for research on various topics including assessment of how textured contact lenses impact the accuracy, employing photometric stereo in iris recognition and PAD, and impact of various NIR illumination on the recognition accuracy.

We compare the proposed method with all iris PAD methods that have available online codes and ones we were able to obtain by contacting the authors of the papers. Extensive experiments on both the \textit{NDIris3D} and a testing partition of \textit{NDCLD'15}~\cite{Doyle2015_NDCLD15} demonstrate that our proposed method has better generalization abilities to unknown textured contact lens types than all existing iris PAD algorithms available as open-source software. 

To summarize, our contributions are the following:
\begin{itemize}
    \item An open-source fusion-based classifier that combines the strengths of 2D \add{(textural features)} and 3D \add{(photometric stereo features)} approaches to \del{address identified weaknesses, and} set\del{s} a new standard of accuracy for open-source iris PAD. \add{The fusion strategy does not arise from the need to remedy their separate robustness issues, but is driven by the fact that 3D and 2D features contain complementary iris PAD-related information.} 
    \item \add{Evaluation of the newly proposed method on two different sensors (IrisGuard AD100 and LG 4000) with completely different illumination setups, to assess the generalization capabilities of the OSPAD-3D compoment of the entire PAD methodology.}
    \item The only systematic evaluation of all (known to us) available open-source iris PAD methods.
    \item A new, large-scale dataset for iris PAD research, with images of \add{the same irises with and without} textured contact lenses of multiple different brands acquired by two sensors and varying NIR illumination setups. \add{This database is useful not only for iris PAD-related research, but also for researching various aspects of matching accuracy deterioration due to wearing contact lenses. The latter is not possible when images of irises with textured contact lens samples do not original from the same subjects who presented their live irises to the sensor. Additionally,} \del{E}\add{e}xperiments show that the same name brand of textured contacts manufactured today does not necessarily have the same texture properties as samples of the same brand from seven years ago~\cite{Doyle2015_NDCLD15}. \add{This introduces a never-considered-before complication to iris PAD development, and we demonstrate in this paper that changes in texture patterns introduced by industry manufacturers have a non-negligible impact on the performance of most iris PAD methods. The offered database should facilitate research on more texture-agnostic iris PAD techniques.}
    \item Experiments to identify subsets of lenses for which the assumptions for 3D method are violated.
\end{itemize}

Along with this paper, both the implementation of our method and the dataset are made publicly available. 

\section{Datasets}

\add{In this section, we introduce the datasets used in this paper. We adopt this unusual ordering of sections because we provide a comparison of all available open-source pre-trained models in Section~\ref{sec:relatedwork} and a development of the fusion idea through empirical evaluation in Section~\ref{sec:method}. With an introduction of the dataset, both sections will be easier to understand.} Two datasets that offer properties we need for both OSPAD-2D and OSPAD-3D methods are used in this work: a subset of the existing \textit{NDCLD’15} dataset and a newly created \textit{NDIris3D} dataset, allowing to evaluate the robustness of the proposed method under varying conditions: time, contact lens brand, sensor, and lens pattern.

\subsection{\textit{NDCLD'15} Dataset}
\label{sec:NDCLD15}
\begin{figure}[!htb]
\centering
\includegraphics[width=0.15\textwidth]{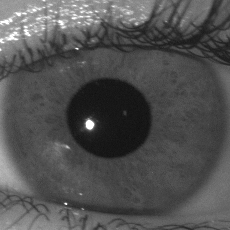}\hfill
\includegraphics[width=0.15\textwidth]{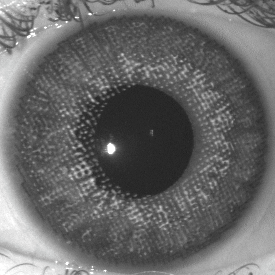}\hfill
\includegraphics[width=0.15\textwidth]{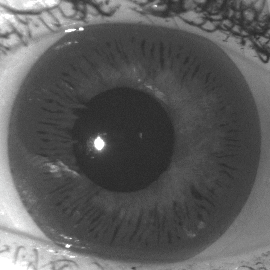}\hfill
\caption{Example iris images with no contact lens (\textbf{left}), with contact lens showing a grid-like (\textit{regular}) texture pattern (\textbf{middle}), and with contact lens without a grid-like texture pattern (\textit{irregular}) (\textbf{right}). The images have their individual brightness adjusted for easy viewing.}
\label{fig:regular_irregular}
\end{figure}

As in the original OSPAD-3D paper~\cite{Czajka_2019_3DPAD}, we use a subset of images from the Notre Dame Contact Lens Detection 2015 (NDCLD’15) dataset, which is the only public dataset known to us at that time that offers iris images (with and without contact lenses) of the same eyes captured with illuminations from two different locations with a short time gap in between \cite{Doyle2015_NDCLD15}. This dataset was collected in 2012, containing a total of 4,068 images (2,664 images of irises wearing textured contact lenses and 1,404 images of irises without any contact lenses) acquired from 119 subjects, from five different brands: Johnson \& Johnson, Ciba Vision, Cooper Vision, Clearlab and United Contact Lens. Among the images of irises wearing textured contact lenses, 1,800 images are of contact lenses with a grid-like texture pattern as shown in Figure~\ref{fig:regular_irregular} (middle), called throughout this paper \textit{regular}, and 864 images are of contact lenses without a grid-like texture pattern as shown in Figure~\ref{fig:regular_irregular} (right), called later \textit{irregular}. There are 37 unique combinations of subject and contact lens brand, because some subjects had images acquired wearing different types of contact lenses and some subjects did not take images with contact lenses. The images are all acquired by the LG4000 iris sensor.

\subsection{\textit{NDIris3D} Dataset}

\begin{figure*}[!ht]
   \centering
   \subfloat[][]{\includegraphics[width=0.15\textwidth]{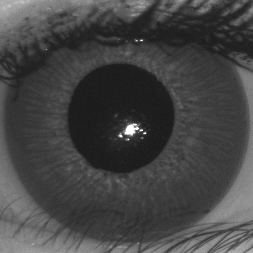}}\quad
   \subfloat[][]{\includegraphics[width=0.15\textwidth]{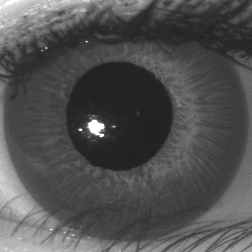}} \quad
   \subfloat[][]{\includegraphics[width=0.15\textwidth]{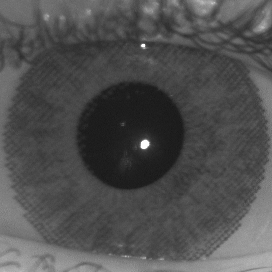}}\quad
   \subfloat[][]{\includegraphics[width=0.15\textwidth]{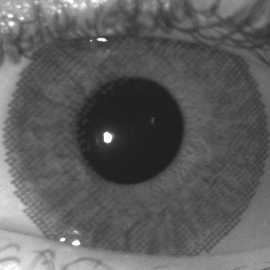}} \quad
   \subfloat[][]{\includegraphics[width=0.15\textwidth]{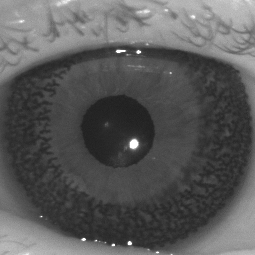}}\quad
   \subfloat[][]{\includegraphics[width=0.15\textwidth]{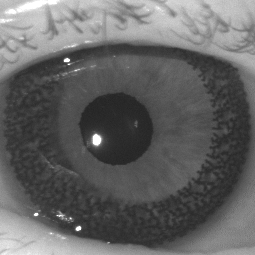}}  \\
   \subfloat[][]{\includegraphics[width=0.15\textwidth]{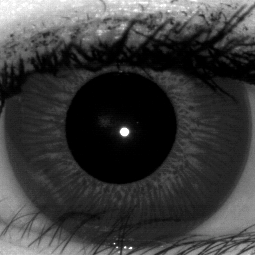}}\quad
   \subfloat[][]{\includegraphics[width=0.15\textwidth]{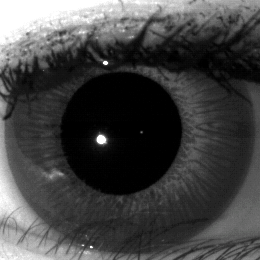}} \quad
   \subfloat[][]{\includegraphics[width=0.15\textwidth]{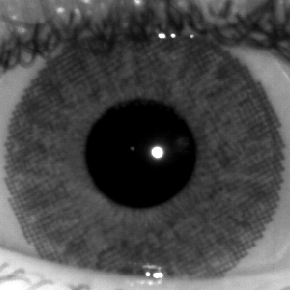}}\quad
   \subfloat[][]{\includegraphics[width=0.15\textwidth]{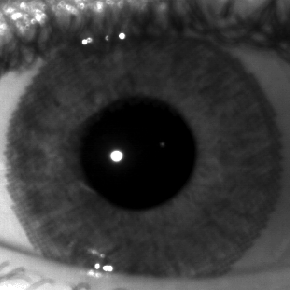}} \quad
   \subfloat[][]{\includegraphics[width=0.15\textwidth]{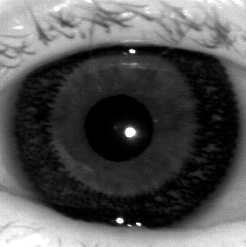}}\quad
   \subfloat[][]{\includegraphics[width=0.15\textwidth]{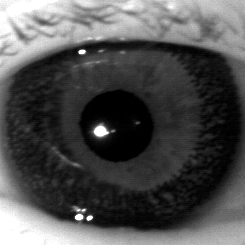}}
   \caption{Examples of image pairs acquired under two different setups of NIR illuminators, as used in OSPAD-3D method. {\bf Upper row:} samples captured by LG4000. {\bf Bottom row:} samples captured by AD100. {\bf First two columns (a,b,g,h):} Johnson \& Johnson. {\bf Second two columns (c,d,i,j):} Ciba Vision. {\bf Last two columns (e,f,k,l):} Bausch\&Lomb. The images acquired by the AD100 sensor have their individual brightness adjusted for easy viewing.}
   \label{fig:AD_LG_paired_examples}
\end{figure*}

To enable comparison of methods from several perspectives, we collected a new dataset \textit{NDIris3D} in 2019. 


In this dataset, images are acquired with 88 subjects (176 irises), each with and without contact lenses (\ie~88 unique combinations of subject and contact lens brand), from three different brands: Johnson \& Johnson, Ciba Vision, and Bausch \& Lomb. The images are collected by both the LG4000 and the AD100 iris sensors under varying near-infrared illumination allowing to design and test photometric stereo-based 3D reconstruction. 
The dataset contains a total of 6,838 images: 3,488 images acquired by LG4000, and 3,362 images acquired by AD100. We denote the subset of images collected by LG4000 as \textit{NDIris3D-LG4000}, and the subset of images collected by AD100 as \textit{NDIris3D-AD100}. Example images from all three brands and both sensors are shown in Figure~\ref{fig:AD_LG_paired_examples}. For LG4000, there are 1,752 images of real irises and 1,736 images of iris wearing textured contact lenses (770 {\it regular} and 966 {\it irregular}); for AD100, there are 1,706 images of real irises and 1,656 images of iris wearing textured contact lenses (742 {\it regular} and 914 {\it irregular}).


\subsection{Changes in Lens Patterns}
\begin{figure}[!htb]
\centering
\subfloat[][]{\includegraphics[width=0.2\textwidth]{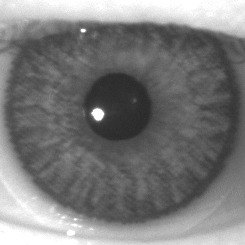}}\quad
\subfloat[][]{\includegraphics[width=0.2\textwidth]{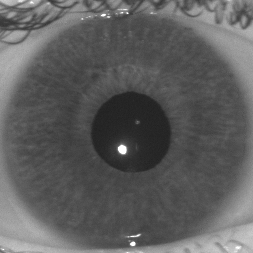}}\\
\subfloat[][]{\includegraphics[width=0.2\textwidth]{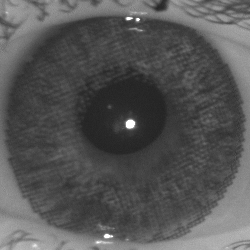}}\quad
\subfloat[][]{\includegraphics[width=0.2\textwidth]{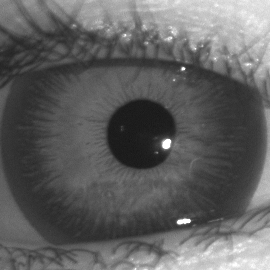}}
\caption{The patterns of contact lenses from the same brands evolved over time. Top left: old Ciba Vision Freshlook Colorblends contact lens; top right: new Ciba Vision Freshlook Colorblends contact lens; bottom left: old Johnson \& Johnson Accuvue contact lens; bottom right: new Johnson \& Johnson Accuvue contact lens.}
\label{fig:old_new_contacts}
\end{figure}

In our study, two of the questions we want to first investigate are (1) whether contact lens manufacturers make changes in the patterns of contact lens over time, and (2) whether these changes influence the accuracy of PAD methods. In the dataset used in~\cite{Czajka_2019_3DPAD}, and in our newly collected dataset there are two brands in common: Ciba Vision and Johnson \& Johnson. Figure~\ref{fig:old_new_contacts} presents a qualitative comparison of the lens patterns of these two brands over time. It is obvious that Johnson \& Johnson Accuvue contact lenses were redesigned, resulting in a clearly observable change in pattern. Compared to the old design, the new contact lens has a thinner ring with a much more irregular inner boundary that no longer follows a circular path. For the Ciba Vision Freshlook contact lenses, it is less clear visible whether there has been a change in lens pattern. This suggests that we may observe changes over time in the contact lenses' appearance, and their strength may be uneven across the brands.


\section{Related Work}\label{sec:relatedwork}
The first iris PAD approach was probably proposed by Daugman~\cite{Daugman_1999_PAD} and employed Fourier analysis to detect artificial patterns in printed contact lenses presented to a sensor. Since then iris PAD has become an increasingly popular research area, and a recent survey~\cite{Czajka_2018_PADsurvey} provides a comprehensive study of the research up to date in iris PAD. Thus, rather than offering yet another survey of iris PAD methods, in this section we focus on PAD aspects important from this work point of view: iris PAD benchmarks and available open-source 2D PAD algorithms that could be used in fusion with our 3D approach.

\subsection{Iris PAD Benchmarks}
In order to evaluate the effectiveness of the state-of-the-art in iris liveness detection, the First International Iris Liveness Detection Competition (LivDet) 2013~\cite{Yambay2014LivDet2013} was held, followed by LivDet 2015~\cite{Yambay2017LivDet2015} and LivDet 2017~\cite{Yambay2017LivDet2017} to further understand the progress of iris PAD methods. Results of these competitions show that iris PAD has improved over time, but also that challenges such as cross-domain detection are under-examined and further advances are needed to better prepare systems for the increase in attack difficulty~\cite{Yambay2019LivDetReview}. The datasets from the competitions, along with the testing protocols, have been made public, aiming to assist progress in iris PAD. However, many of the methods either participated anonymously or have no implementations available to other researchers, which makes the replication and comparison of results presented in papers difficult or impossible. To this end, we conduct a study to systematically evaluate all currently and publicly available open source iris PAD implementations.

\subsection{Open Source Iris PAD Methods}
\label{sec:opensource_overview}
To the best of our knowledge, we collected all the iris PAD methods that have codes available. At the time of this paper, there was one method with code available online~\cite{McGrath_2018_OSPAD}. We contacted the corresponding authors of 12 iris PAD papers and eventually obtained working implementations presented in three additional papers~\cite{Hu_2016_RegionalFeatures, Gragnaniello_2016_IrisSclera, Gragnaniello_2016_DACNN}. Here, we provide a brief overview of all these four methods.

Hu \cite{Hu_2016_RegionalFeatures} performs PAD by designing regional features based on feature distribution in neighboring regions. The regional features are constructed based on spatial pyramid and relational measure. The spatial pyramid extracts features on multiple levels of resolution, and the relational measure performs convolution on features with variable-size kernels. For low-level features that build the models, the authors investigate a variety of feature extractors, including Local Binary Patterns (LBP), Local Phase Quantization (LPQ), and intensity correlogram. In the rest of this paper, we will refer to this method as \textit{RegionalPAD}. \par

Gragnaniello \cite{Gragnaniello_2016_IrisSclera} uses both iris and sclera regions for feature extraction, arguing that the sclera region also contains rich information about iris liveness. Scale-invariant local descriptors (SID) are applied on the segmented iris and sclera areas. Then, the features are summarized through a bag-of-feature method and classified by a linear Support Vector Machine (SVM). In the rest of the paper, we will refer to this method as \textit{SIDPAD}. \par

Seeing the effectiveness of Convolutional Neural Networks (CNN) in many other areas of image processing, Gragnaniello \cite{Gragnaniello_2016_DACNN} proposed a CNN architecture for biometric spoofing detection. The authors incorporate domain-specific knowledge into the design of the network architecture and the loss functions. The method could be applied to PAD in multiple biometric modes, including face and iris. In the rest of the paper, we will refer to this method as \textit{DACNN}. \par

McGrath \etal \cite{McGrath_2018_OSPAD} proposed an open source PAD method, implemented in both C++ and Python using only open source resources, to classify the iris image as either an authentic iris or a textured contact lens. As an extension of \cite{Doyle2015_NDCLD15}, this method employs multi-scale Binary Statistical Image Features (BSIF) as the feature extractor and uses an ensemble of multiple classifiers, including SVM, Multilayer Perceptrons (MLP), and Random Forests (RF). This method achieves state-of-the-art performance, obtaining an accuracy on LivDet-Iris 2017 on par with that of the LivDet-Iris 2017 winner. Note that this method does not employ any iris image segmentation in its pre-processing. Instead, the authors leverage the fact that commercial iris sensors tend to acquire images where the iris is more or less centered and use the best guess (a box centered in the image) of the iris' position in the input image. If an open-source segmentation software were used, the overall method could achieve even higher accuracy while still being open-source. In the rest of the paper, we will refer to this method as \textit{OSPAD-2D}.

In this paper, we systematically investigate the above four iris PAD methods, along with our previous work and the proposed method in this paper, under various testing conditions (lens brands, sensor, pattern type), and propose a fusion-based method that achieves robust performance under all testing conditions. It is important to note that our OSPAD-3D needs a pair of images to make predictions, while the other four methods work with single images. To make a fair comparison and to make fusion possible, we form each sample as an image pair. For methods other than OSPAD-3D, score level fusion is performed for the two images in the pair to produce a final decision for that sample. The question we are answering here is: if we have two iris images, instead of one, captured by state-of-the-art commercial sensors in a way allowing to extract 3D information about the observed object and not making the acquisition harder or longer, do the accuracy and generalization capabilities of the PAD grow?

\subsection{Performance of Available Open-Source Methods with Pre-trained Models}

In addition to the source codes, we collected all pre-trained models available at the time of preparing this paper. Such models were available only for methods published in~\cite{McGrath_2018_OSPAD} (OSPAD-2D) and~\cite{Hu_2016_RegionalFeatures} (RegionalPAD). The authors of OSPAD-2D suggested three sets of BSIF filter parameters selected from NDCLD'15, Clarkson, and IIITD subsets of LivDet-Iris 2017 benchmark. We denote methods using these three sets of filters as \textit{OSPAD-2D-ND}, \textit{OSPAD-2D-Clarkson}, and \textit{OSPAD-2D-IIITD}, respectively.

In this paper, we follow the ISO/IEC 30107-3:2017 and use the following PAD error metrics, widely deployed in biometric performance evaluation efforts, including LivDet-Iris 2017~\cite{Yambay2017LivDet2017}: 

\begin{itemize}
    \item \textit{Attack Presentation Classification Error Rate} (APCER), which is the proportion of {\it attack presentations} incorrectly classified as {\it bona fide presentations}, and
    \item \textit{Bona Fide Presentation Classification Error Rate} (BPCER), which is the proportion of {\it bona fide presentations} incorrectly classified as {\it presentation attacks}.
\end{itemize}

Also, when the term {\it Accuracy} is used, it refers to the fraction of correct classifications (either spoofed or authentic samples) to the total number of classifications made.

The methods' performance on the \textit{NDIris3D} dataset is shown in Table~\ref{table:pretrained}. From these results, pre-trained models provided by \textit{RegionalPAD} fail to capture the real and fake distributions of \textit{NDIris3D} from both LG4000 and AD100 sensors, rejecting a large number images of authentic irises. In comparison, the three versions of \textit{OSPAD-2D} perform slightly better on AD100 data and a lot better on LG4000 data. Among the three sets of filters, the one developed on Clarkson subset, \textit{OSPAD-2D-Clarkson}, performs the best for both AD100 and LG4000, and we will keep this set of filters for the rest of our experiments with the \textit{OSPAD-2D} method (we skip ``Clarkson'' and keep ``OSPAD-2D'' for brevity later in the paper). We conjecture that this accuracy gap between \textit{OSPAD-2D} and \textit{RegionalPAD} comes from the fact that there is an overlap in contact brands (but {\it not} samples) between the training data of \textit{OSPAD-2D} (\ie~\textit{NDCLD'15}) and \textit{NDIris3D}.

\begin{table*}
\begin{center}
\caption{Performance of all available open-source iris PAD methods, using pre-trained models from their authors, on the new \textit{NDIris3D} dataset.}
\label{table:pretrained} 
\begin{tabular}{ccccccc}
\toprule
\multirow{3}{*}{Methods} & \multicolumn{6}{c}{Subset of NDIris3D}\\
 & \multicolumn{3}{c}{NDIris3D-LG4000} & \multicolumn{3}{c}{NDIris3D-AD100} \\
\cline{2-7}
 & Accuracy (\%) & APCER (\%) & BPCER (\%) & Accuracy (\%) & APCER (\%) & BPCER (\%) \\
\midrule
OSPAD-2D ND & 77.75 & 33.81 & 10.79 & 58.89 & 77.85 & 4.67 \\ 
OSPAD-2D Clarkson & 85.40 & 29.09 & 0.23 & 67.46 & 65.32 & 0.04 \\ 
OSPAD-2D IIITD & 79.96 & 39.98 & 0.29 & 66.75 & 66.63 & 0.15\\ 
RegionalPAD & 50.53 & 0.11 & 98.9 & 53.25 & 3.51 & 89.63 \\ 
\bottomrule
\end{tabular}
\end{center}
\end{table*}


\section{Method}\label{sec:method}

\subsection{OSPAD-3D}

While all other open source methods mentioned in Section~\ref{sec:opensource_overview} learn to determine the decision boundary based on the textural differences between authentic irises and those with textured contact lenses, OSPAD-3D is based on 3D features estimated by photometric stereo from two iris images taken in near-infrared illumination coming from two different locations \cite{Czajka_2019_3DPAD}. We provide a brief description of OSPAD-3D here. 

\subsubsection{General Photometric Stereo Approach}
Photometric stereo is a computer vision method of estimating the surface normal vectors by observing an object
under illuminations from different directions by a single
fixed-position camera. Assuming we use $k$ point-wise illuminators that generate $k$ Lambertian reflections (\ie~diverging almost equally in all directions) from each point of
a surface with uniform albedo, we can use a linear model
binding these quantities:

\begin{equation}
\mathbf{I} = \mathbf{L}\mathbf{\hat{n}} = \mathbf{L}c\mathbf{n}
\label{eqn:i}
\end{equation}

\noindent
where $\mathbf{I}$ is a vector of the observed $k$ intensities, $\mathbf{L}$ is a $3 \times k$ matrix of $k$ known light directions, $c$ represents a uniform albedo, and $\mathbf{n}$ is the surface unit normal vector to be estimated. This yields

\begin{equation}
\mathbf{\hat{n}} = 
\begin{cases}
    \mathbf{L}^{-1} \mathbf{I} & \text{if} \; k = 3 \\
    (\mathbf{L}^{T} \mathbf{L})^{-1}\mathbf{L}^{T}\mathbf{I} & \text{if} \; k \neq 3
\end{cases}
\label{eqn:nhat}
\end{equation}

\noindent
where $(\mathbf{L}^{T} \mathbf{L})^{-1}\mathbf{L}^{T}$ is  Moore-Penrose pseudoinverse of $\mathbf{L}$. Assuming the albedo $c$ is uniform for all  points, we can estimate the unit surface normals as

\begin{equation}
\mathbf{n} = \frac{\mathbf{\hat{n}}}{\|\mathbf{\hat{n}\|}}
\label{eqn:n}
\end{equation}

\noindent
where $\|x\|$ is the $\ell^2$ (Euclidean) norm of $x$. Since we need to find two unknowns in $\mathbf{\hat{n}}$, $k=2$ images taken under two different lighting conditions are necessary to solve the equation (\ref{eqn:n}) and calculate $\mathbf{n}$. In our formulation of the PAD problem, only two images are required. 

The normal vectors $\mathbf{n}$ are estimated for each picture point $(x,y)$, so $\mathbf{I}_{x,y}$ and $\mathbf{n}_{x,y}$ should be used for the observed intensities and normal vector at point $(x,y)$, respectively. We skipped $(x,y)$ subscript in equations (\ref{eqn:i} - \ref{eqn:n}) for clarity.

\subsubsection{OSPAD-3D pipeline}

\begin{figure}[!htb]
\centering
\includegraphics[width=0.46\textwidth]{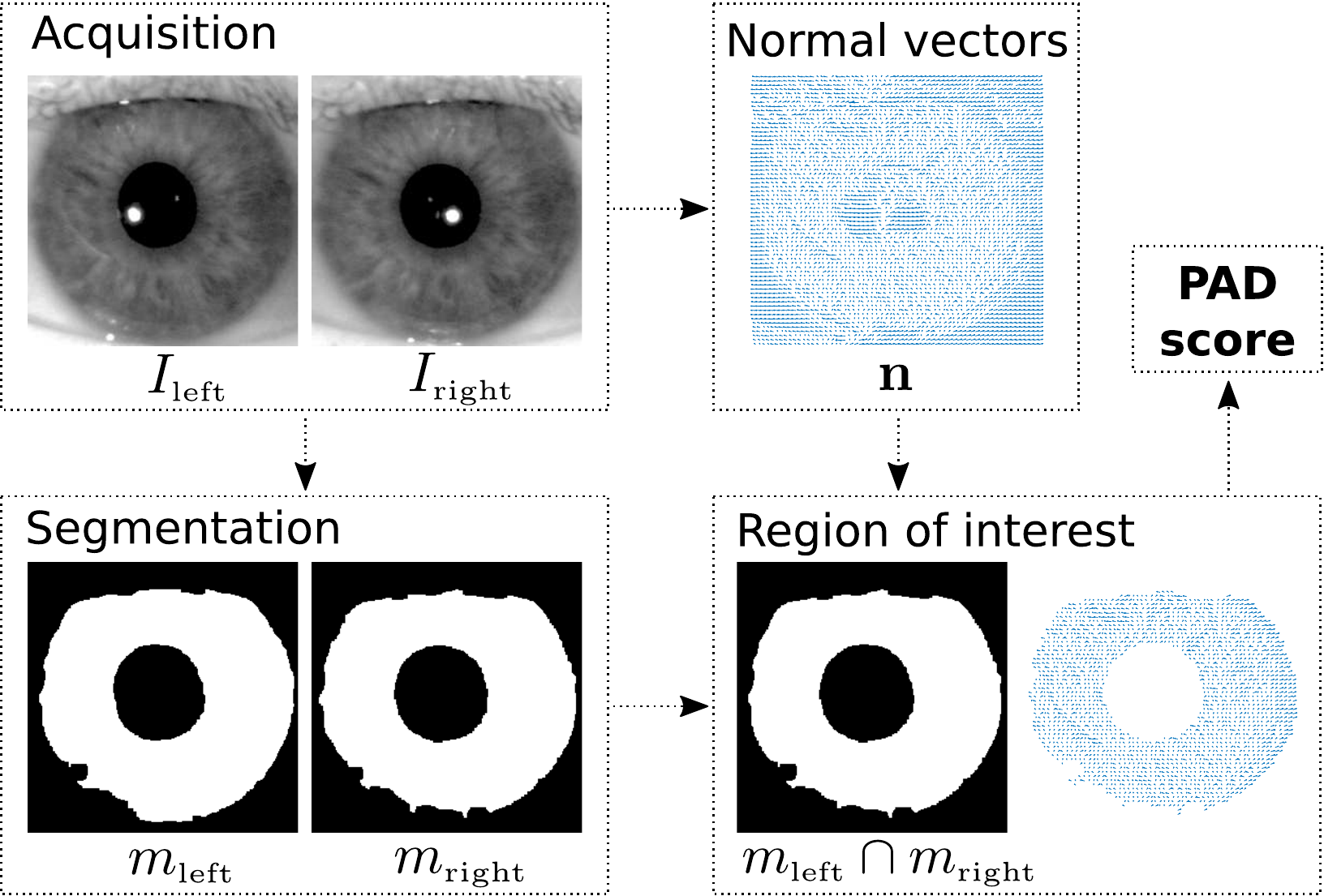}
\caption{Components of the OSPAD-3D pipeline.}
\label{fig:method}
\end{figure}

The overall pipeline of OSPAD-3D is shown in Figure~\ref{fig:method}. In our implementation, we use two observed intensities in equation (\ref{eqn:i}) for each picture point $(x,y)$:

$$
\mathbf{I}_{x,y} = 
\begin{bmatrix}
 	I_{\mbox{\tiny left}}(x,y)\\
 	I_{\mbox{\tiny right}}(x,y)
 \end{bmatrix}
$$

\begin{figure*}[!htb]
\centering
\subfloat[][]{\includegraphics[width=0.51\textwidth]{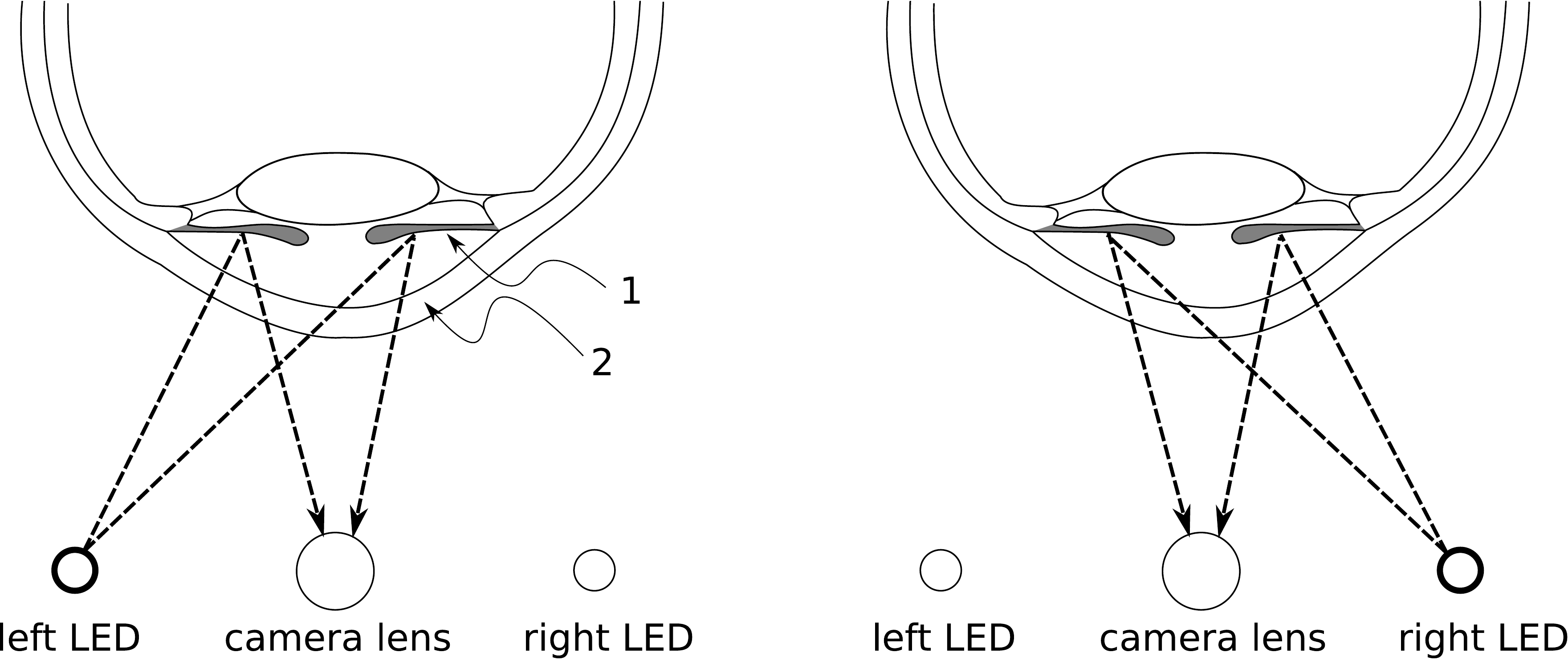}}\quad
\subfloat[][]{\includegraphics[width=0.22\textwidth]{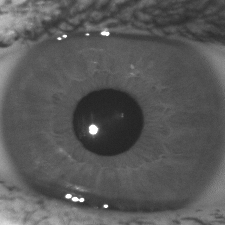} \includegraphics[width=0.22\textwidth]{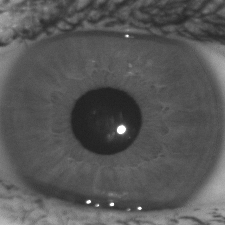}}\quad
\caption{a) In case of observing an eye not wearing a textured contact lens, or wearing a transparent contact lens, the NIR light rays go through the cornea (2) and are reflected from the iris (1); b) The corresponding $I_{\mbox{\tiny left}}$ and $I_{\mbox{\tiny right}}$ iris images. Differences in shadows observed in these two pictures are small, which ends up with a reconstruction of a roughly planar surface.}
\label{fig:e1}
\end{figure*}

Figure \ref{fig:e1} (a) illustrates a typical setup of two near-infrared illuminators placed equidistant to the camera lens. This allows to generate a pair of iris images as shown in Fig. \ref{fig:e1} (b). The external surface (visible to us) of the iris can be considered as a more Lambertian than specular surface. So, using either left or right illuminator produces very similar iris images. Certainly, the iris surface is not perfectly flat and this should manifest in different shadows visible in the left and right images. However, the resolution of commercial iris recognition sensors compliant to ISO/IEC 19794-6 is rather small (normally approx. 200 pixels across iris diameter are used, with 120 pixels being the standard requirement) when compared to the size of three-dimensional objects such as crypts, and hence the observed differences between $I_{\mbox{\tiny left}}$ and $I_{\mbox{\tiny right}}$ are small. The photometric stereo method will end up with estimation of normal vectors that should not differ too much from the average normal vector estimated for this object.

\begin{figure*}[!htb]
\centering
\subfloat[][]{\includegraphics[width=0.51\textwidth]{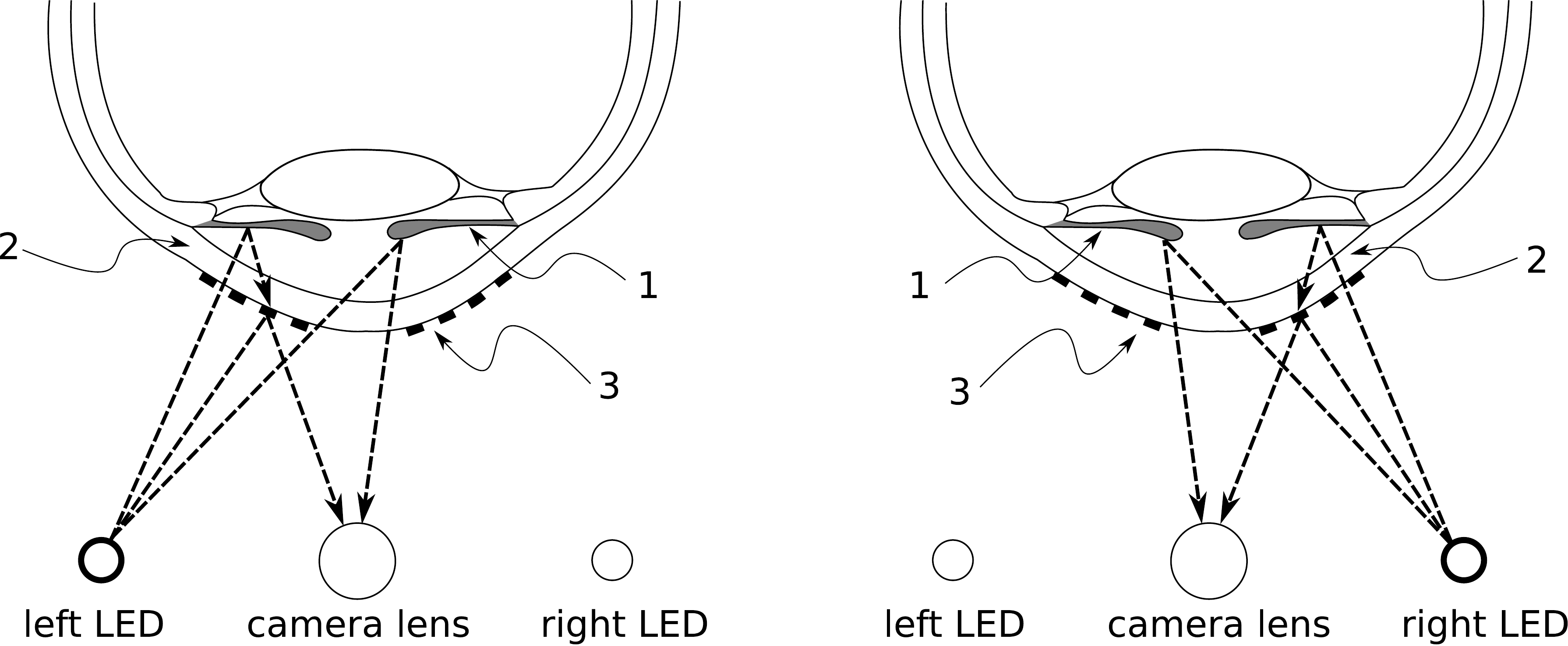}}\quad
\subfloat[][]{\includegraphics[width=0.22\textwidth]{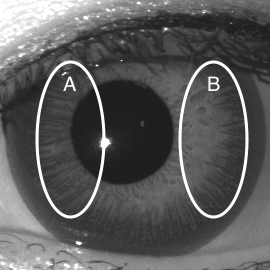} \includegraphics[width=0.22\textwidth]{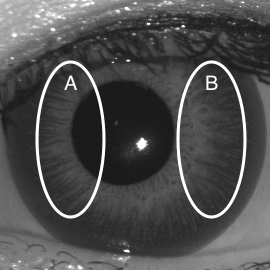}}\quad
\caption{a) In case of observing an eye wearing a textured contact lens (3), the NIR light rays either go through the cornea (2) and, reflected from the iris, are registered by the camera, or they are reflected by the textured contact lens, or they are reflected from the iris but blocked by the opaque texture printed on the lens. b) The corresponding $I_{\mbox{\tiny left}}$ and $I_{\mbox{\tiny right}}$ iris images. Note that significant differences in generated shadows between the left- and right-illuminated iris with textured contact lens, especially in areas marked as A and B.}
\label{fig:e2}
\end{figure*}

Figure \ref{fig:e2} illustrates identical capture procedure (a) and the resulting images (b) when an eye wearing textured contact lens is imaged. One should note shadows made by \del{the }the partially opaque texture printed on the lens and observed in different places, depending on which illuminator was used to illuminate the object. Except for large shadows observed in regions marked as (A) and (B), we can also see differences how the printed texture generates image features under illumination at different angles. Consequently, for this object the photometric stereo will end up with highly variable normal vectors due to irregular and noisy surface that is being estimated.

\begin{figure}[!htb]
\centering
\subfloat[][]{\includegraphics[width=0.22\textwidth]{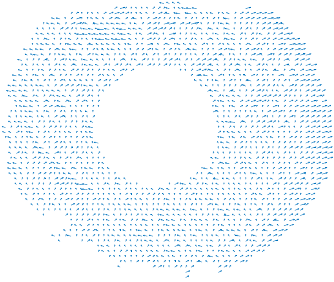}}\quad
\subfloat[][]{\includegraphics[width=0.22\textwidth]{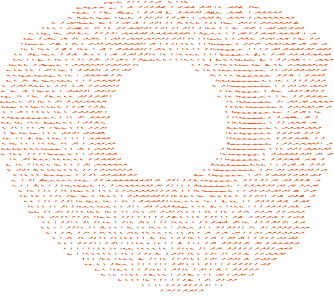}}
\caption{Quiver plots depicting estimated normal vectors for authentic iris (a) and iris with textured contact lens (b).}
\label{fig:quiver}
\end{figure}

The normal vectors estimated for the images shown in Figs. \ref{fig:e1} (b) and \ref{fig:e2} (b) are illustrated as quiver plots in Fig. \ref{fig:quiver} (a) and \ref{fig:quiver} (b), respectively. Note a higher variability of the estimated normal vectors for an eye wearing a textured contact lens. Note also that we do not consider normal vectors outside the iris annulus, and for portions of the iris occluded by eyelids and eyelashes. This is accomplished by calculating the occlusion masks $m_{\mbox{\tiny left}}$ and $m_{\mbox{\tiny right}}$ corresponding to $I_{\mbox{\tiny left}}$ and $I_{\mbox{\tiny right}}$ iris images. For non-occluded iris pixels $m=1$, and for background pixels $m=0$.

Let $\mathbf{\bar{n}}$ be the average normal vector within the non-occluded iris area. The Euclidean distances between normals and their average are smaller for an approximately flat iris when compared to an irregular object composed with an iris and a textured contact lens. Consequently, in this presentation attack detection method the variance of an Euclidean distance between the normals and their average, calculated in non-occluded iris area, is used as the PAD score:

\begin{equation}
\label{eqn:score}
q = var \| \mathbf{n}_{x,y} - \mathbf{\bar{n}} \|
\end{equation}

\noindent
where $\|x\|$ is the $\ell^2$ (Euclidean) norm of $x$, and

$$
\mathbf{\bar{n}} = \frac{1}{N}\sum_{x,y} \mathbf{n}_{x,y},
$$

\noindent
where $N$ is the number of non-occluded iris points, and
$$
\{(x,y): m_{\mbox{\tiny left}}(x,y) \cap m_{\mbox{\tiny right}}(x,y) = 1\}.
$$

We expect to observe a larger variance calculated by equation (\ref{eqn:score}) for irises wearing textured contact lenses than for irises not wearing textured contacts, or wearing transparent contacts.

In our original work~\cite{Czajka_2019_3DPAD}, the iris regions are segmented by OSIRIS~\cite{Othman_2016_OSIRIS}. In this work, to make our method completely open-source and  to obtain more accurate segmentations for irregular shapes, we adopt a SegNet-based segmentation~\cite{Trokielewicz_2018_IrisSegNet}. 


\subsection{OSPAD-3D and Violations of Its Assumptions}


\begin{table}
\begin{center}
\caption{Accuracy of OSPAD-3D on different brands from \textit{NDIris3D}.}
\label{table:brand_error} 
\begin{tabular}{ccc}
\toprule
\multirow{2}{*}{Brand} & \multicolumn{2}{c}{\del{Acc.} \add{1-APCER} (\%)}\\ \cline{2-3}
 & LG4000 & \add{AD100} \\ 
\midrule
Johsnon\&Johnson & $60.16$ & \add{$56.59$}\\ 
Ciba Vision & $80.21$ & \add{$58.55$} \\ 
Bausch\&Lomb & $98.97$ & \add{$97.70$} \\ 
\bottomrule
\end{tabular}
\end{center}
\end{table}

OSPAD-3D has been proven to have high accuracy and good generalization abilities when unseen contact lens patterns are present in the testing set \cite{Czajka_2019_3DPAD}. However, due to dynamics of the contact lens manufacturing that follows market trends (not necessarily correlated in any way with potential usefulness of such lens in biometric presentation attacks), there are circumstances where this method offers limited accuracy. In the following experiments we show that the assumptions about shadows in OSPAD-3D break down for contact lenses whose patterns are highly opaque (original iris patterns are hardly visible when covered by the contact lens). When the lens has a highly opaque pattern, the differences in shadows are not well pronounced when the iris is illuminated from different direction, and therefore the reconstructed surface is relatively on the flat side, making OSPAD-3D classify the sample as an authentic iris. Here, we demonstrate this phenomenon on the newly collected \textit{NDIris3D}, with the PAD score threshold determined from a combined dataset from \textit{NDCLD’15} and the dataset from LivDet-Iris 2017. The accuracy of OSPAD-3D on each brand of lenses are shown in Table~\ref{table:brand_error}. It can be clearly observed that OSPAD-3D oftentimes fails for Johnson\&Johnson contact lenses \add{and Ciba Vision contact lenses}\del{whose patterns are more opaque compared to Ciba Vision and Bausch\&Lomb}. \add{This is because these contact lenses are more opaque and have larger coverage of the iris and a less clear inner border, as shown in Figure~\ref{fig:AD_LG_paired_examples}, while Bausch\&Lomb has a clear-cut inner boundary and has a porous texture. Note that the accuracy of OSPAD-3D generally dropped for images taken with the AD100 sensor. This is because no images in the training set are taken with the AD100 sensor, making this test a cross-sensor experiment.} Through these experiments, we see an important real-world complication for spoofing detection: it seems that a method successful in detecting contact lens attacks may ``age'' and be less effective due to fluctuations in contact lens manufacturing process over time.


\subsection{Evaluation of Open-Source Methods with Fine-Tuned Models}
\label{sec:open_source}

We have thus conducted a series of experiments to study the impact of change in lens pattern over time on the accuracy of the state-of-the-art iris PAD methods. First, we test the accuracy of iris PAD methods when the changes in lens patterns are minimal between the training and testing subsets. \del{We split \textit{NDCLD'15} into two subject-disjoint halves, trained all the methods on one of the two halves and tested on the other half. Table~\ref{table:old_old}
shows the average accuracy for the two-way half-half experiments. We repeat this for \textit{NDIris3D-LG4000} as well and the results are shown in Table~\ref{table:new_new}. In general, all the methods achieved satisfactory accuracy for both tests. The best-performing method on average is OSPAD-2D, followed by RegionalPAD and OSPAD-3D.} \add{On \textit{NDCLD'15}, we perform a five-fold cross-validation with subject-disjoint splits. Table~\ref{table:old_old} shows the average accuracy with standard deviation in the parentheses. We repeat this experiment for \textit{NDIris3D-LG4000} as well and the results are shown in Table~\ref{table:new_new}. All methods show solid performance on both datasets. The best-performing method is OSPAD-2D, with near-perfect performance on both datasets. This shows that OSPAD-2D captures the real and attack distributions very well. Also, note that OSPAD-3D achieves relatively low accuracy on both datasets.} This is reasonable because OSPAD-3D does not leverage the textural information of the contact lenses, and therefore it does not have any advantages when the lens patterns in the testing set are seen in the training set. \add{Therefore, the results of these experiments indicate that testing scenarios where the lens patterns are seen previously are not very challenging, as most recent benchmark methods are all able to perform robustly.} \del{ Surprisingly, DACNN and SIDPAD also learn the decision boundary between authentic irises and contact lenses based on their textural difference, but both methods only performed well on the new dataset while accepting many attacks as authentic irises on the old dataset. We conjecture that this is because of their sensitivity to the dataset split, failing to capture the distribution of different datasets sufficiently.}


\begin{table}
\begin{center}
\caption{\add{Performance of methods in subject-disjoint five-fold cross-validation using \textit{NDCLD'15}}. Properties of the contact lens pattern are similar in train and test partitions. Both average and standard deviation are reported.}
\label{table:old_old}
\begin{tabular}{c>{\color{myBlue}}c>{\color{myBlue}}c>{\color{myBlue}}c}
\toprule
\multirow{2}{*}{Methods} & \multicolumn{3}{c}{Performance}\\ \cline{2-4}
 & Acc. (\%) & APCER (\%) & BPCER (\%)\\ 
\midrule
OSPAD-3D & $92.14~(\pm 6.31)$ & $7.14~(\pm 3.50)$ & $8.57~(\pm 12.12)$ \\ 
OSPAD-2D & $97.32~(\pm 3.97)$ & $5.36~(\pm 7.94)$ & $0.00~(\pm 0.00)$ \\ 
DACNN & $93.33~(\pm 2.34)$ & $5.18~(\pm 5.06)$ & $8.21~(\pm 8.42)$ \\ 
SIDPAD & $91.99~(\pm 4.79)$ & $13.83~(\pm 10.39)$ & $2.32~(\pm 0.90)$ \\ 
RegionalPAD & $94.73~(\pm 3.01)$ & $8.93~(\pm 7.13)$ & $1.61~(\pm 1.88)$ \\ 
\bottomrule
\end{tabular}
\end{center}
\end{table}

\begin{table}
\begin{center}
\caption{Same as in Table. \ref{table:old_old}, except the newly collected \textit{NDIris3D-LG4000} dataset was used.}
\label{table:new_new} 
\begin{tabular}{c>{\color{myBlue}}c>{\color{myBlue}}c>{\color{myBlue}}c}
\toprule
\multirow{2}{*}{Methods} & \multicolumn{3}{c}{Performance}\\ \cline{2-4}
 & Acc. (\%) & APCER (\%) & BPCER (\%)\\ 
\midrule
OSPAD-3D & $93.21~(\pm 0.61)$ & $7.36~(\pm 2.60)$ & $6.21~(\pm 1.70)$ \\ 
OSPAD-2D & $99.54~(\pm 0.39)$ & $0.92~(\pm 1.36)$ & $0.92~(\pm 1.10)$ \\ 
DACNN & $98.01~(\pm 0.63)$ & $1.55~(\pm 1.64)$ & $2.49~(\pm 1.50)$ \\ 
SIDPAD & $95.99~(\pm 1.33)$ & $5.00~(\pm 2.56)$ & $3.03~(\pm 1.22)$ \\ 
RegionalPAD & $94.79~(\pm 2.59)$ & $5.26~(\pm 3.16)$ & $5.16~(\pm 3.16)$ \\ 
\bottomrule
\end{tabular}
\end{center}
\end{table}



Next, we evaluate the accuracy of iris PAD when the train and test subsets contain contact lenses from the same manufacturers but obtained several years apart. This is the first such evaluation known to us. We want to investigate whether changes in contact lens pattern designs introduced by the manufacturers over time have an impact on the performance of the algorithms. We train the methods on Ciba Vision and Johnson\&Johnson data (with the same number of real iris samples as contact lens samples) from \textit{NDCLD’15} and test on Ciba Vision and Johnson\&Johnson data (again, with the same number of real and contact lens iris samples) from \textit{NDIris3D-LG4000}. As shown in Table~\ref{table:brand_oldvsnew}, we observe a drop in accuracy for all the methods, and especially in the methods based on textural information. OSPAD-3D now performs the best, which is reasonable because it does not rely on the texture of contact lenses to make decisions. The four other methods obtain similar performance, failing to generalize well on new lens patterns. This is an important observation, since changes made by manufacturers are unpredictable. This implies that for iris PAD methods to maintain accuracy, they need to be constantly updated to include training samples of the latest contact lenses, even from the same manufacturers.

\begin{table}
\begin{center}
\caption{Accuracy of methods when trained on Ciba Vision and Johnson\&Johnson samples from the \textit{NDCLD’15} dataset, and tested on Ciba Vision and Johnson\&Johnson samples from the \textit{NDIris3D-LG4000} dataset.}
\label{table:brand_oldvsnew} 
\begin{tabular}{cccc}
\toprule
\multirow{2}{*}{Methods} & \multicolumn{3}{c}{Performance}\\ \cline{2-4}
 & Acc. (\%) & APCER (\%) & BPCER (\%)\\ 
\midrule
OSPAD-3D & $82.74$ & $31.64$ & $3.08$ \\ 
OSPAD-2D & $74.80$ & $51.91$ & $0$ \\ 
DACNN & $70.07$ & $43.29$ & $16.75$ \\ 
SIDPAD & $74.31$ & $48.18$ & $3.53$ \\ 
RegionalPAD & $72.24$ & $52.80$ & $3.08$ \\ 
\bottomrule
\end{tabular}
\end{center}
\end{table}

\subsection{2D-3D Fusion}
\label{sec:2D_3D_Fusion}

\begin{figure}
    \centering
    \includegraphics[width=0.48\textwidth]{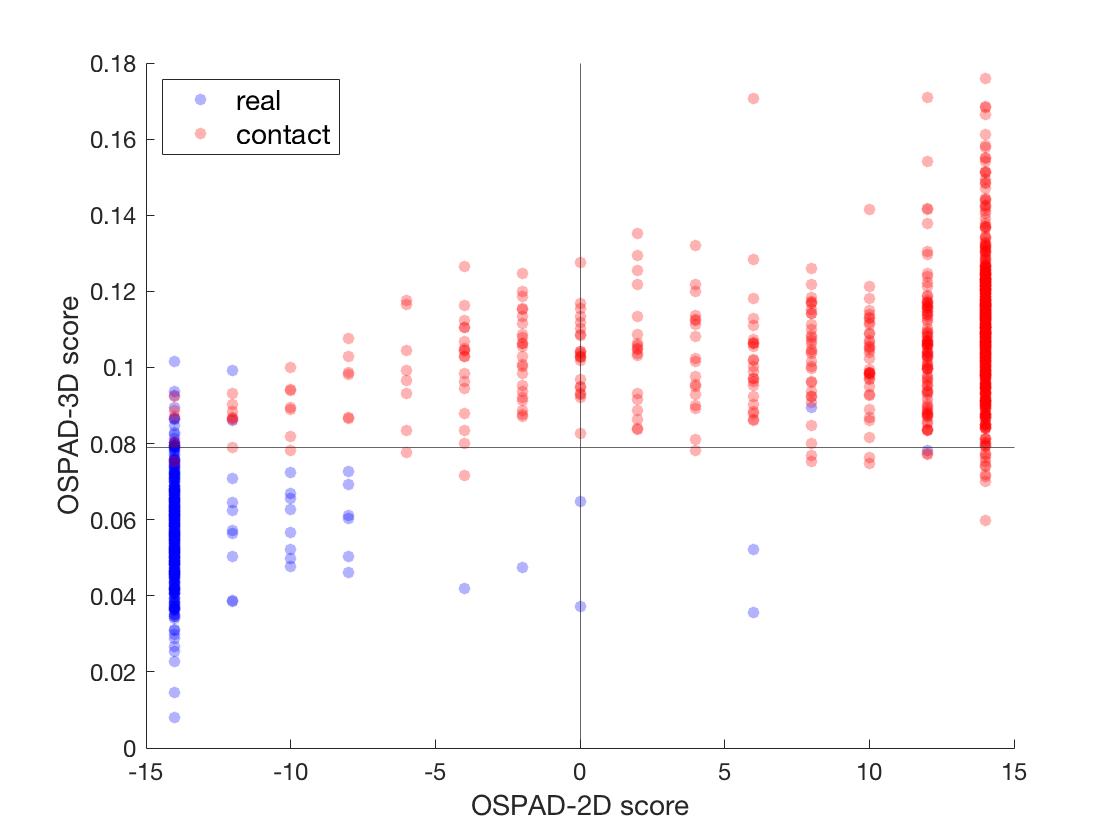}
    \caption{PAD score distributions of the OSPAD-3D method vs. the OSPAD-2D method.}
    \label{fig:fusion_3D_vs_2D}
\end{figure}

From Tables~\ref{table:old_old},~\ref{table:new_new}, and~\ref{table:brand_oldvsnew} we see that some texture description-based methods are more accurate when making predictions on samples similar to the training examples, while the OSPAD-3D is more robust when tested against samples with unknown patterns. Therefore, it is a natural choice to fuse these two types of approach. We propose to fuse OSPAD-3D with OSPAD-2D, as OSPAD-2D has the highest overall accuracy among all the 2D methods. To make a fair comparison with other PAD methods, we design the fusion rules based solely on the \textit{NDCLD15 LG4000} subset. We perform a random, subject-disjoint, lens-pattern-disjoint split to obtain a (60\%, 40\%) \del{train-validation} \add{train-test} split. Figure~\ref{fig:fusion_3D_vs_2D} shows the scatter plot of PAD scores, where the coordinates are (OSPAD-2D score, OSPAD-3D score). We observe that OSPAD-2D has a high APCER and a low BPCER. Leveraging this characteristic, we design the cascaded fusion algorithm denoted as OSPAD-fusion:
\begin{itemize}
    \item for samples predicted as contact lenses by OSPAD-2D, predict them as contact lenses,
    \item for all other samples, use the OSPAD-3D as final predictions.
\end{itemize}




The fusion results are presented in Table~\ref{table:fusionpad_345}. While the fusion rules are simple, they are effective when tested in the same scenarios as shown in Tables~\ref{table:old_old},~\ref{table:new_new}, and~\ref{table:brand_oldvsnew}. When trained on half of the \textit{NDCLD’15} dataset and tested on the other half of the same dataset, a clear performance boost is obtained by fusing the two methods, achieving a performance on par with the best performing RegionalPAD. When trained on half of the \textit{NDIris3D} dataset and tested on the other half of the same database, the fusion performance is still satisfactory, without being affected by OSPAD-3D. When trained on the old Ciba Vision and Johnson\&Johnson samples (taken from \textit{NDCLD’15}) and tested on the new Ciba Vision and Johnson\&Johnson samples (taken from \textit{NDIris3D}), we obtain a significant performance gain: 8.99\% over OSPAD-3D and 20.56\% over OSPAD-2D. This provides evidence of the effectiveness of the fusion, leveraging the variety in errors made by OSPAD-2D and OSPAD-3D and avoiding the majority of both.


\begin{table}
\begin{center}
\caption{Accuracy of the {\it OSPAD-fusion} method in the three previous experimental setups.}
\label{table:fusionpad_345} 
\begin{tabular}{cccc}
\toprule
\multirow{2}{*}{Experimental Setup} & \multicolumn{3}{c}{Performance}\\ 
\cline{2-4}
 & Accuracy (\%) & APCER (\%) & BPCER (\%)\\ 
\midrule
Table~\ref{table:old_old} & $94.10$ & $0.69$ & $11.11$ \\ 
Table~\ref{table:new_new} & $96.12$ & $2.08$ & $5.66$ \\ 
Table~\ref{table:brand_oldvsnew} & $90.18$ & $13.02$ & $6.68$ \\ 
\bottomrule
\end{tabular}
\end{center}
\end{table}

While the inferred fusion rules are shown to be effective through experiments, we are aware that there are several other possible fusion schemes:
\begin{enumerate}
    \item Score-level fusion of OSPAD-3D and OSPAD-2D after normalizing their scores.
    \item Cascaded fusion with OSPAD-3D being the first classifier and OSPAD-2D being the second.
    \item Majority voting with multiple OSPAD-2D classifiers with different parameters and OSPAD-3D.
    \item Majority voting with OSPAD-3D and other 2D-based PAD methods.
\end{enumerate}
We tested the performance of the above fusion schemes on the same aforementioned experimental setup, but none of them provides an accuracy boost larger than OSPAD-fusion.


\add{
\subsection{Notes on data splits into train, validation and test subsets}
In machine learning, we typically speak of ``training-set'' (used to create a model), ``validation-set'' (to score a trained model and {\it project} its performance on unseen data) and ``test set'' (to provide an {\it actual} performance on unseen data). It was of paramount importance to not use any information from the test set in any final evaluations presented in Sections \ref{sec:open_source} and \ref{sec:2D_3D_Fusion}. Specifically, the datasets are split into subject-disjoint halves and neither of the halves is used to re-design the methods. That is: (a) for traditional vision-based methods, the ``training set'' is used to train the SVM (in the cases of OSPAD-2D, SIDPAD and RegionalPAD methods), or setting the threshold (in the case of the OSPAD-3D method); (b) for deep learning-based method (DACNN), the ``training set'' is further split into actual ``training'' partition (to train the neural networks whose architectures are fixed) and ``validation'' partition (to decide when to stop the training). Parts denoted as ``test'' in this manuscript were never used in any development to offer final and fair evaluation.}
\section{Experiments with OSPAD-fusion and Results}

\subsection{Experimental Setup}
To evaluate the performance of the methods, we conduct experiments under two testing scenarios:

\begin{enumerate}
\item {\it train old, test new:} In order to understand how well the methods generalize for a different sample distribution (different subjects, a 7-year time interval, and changes in lens patterns) and how robust they are when the sensor is changed in the testing set, we train on the entire \textit{NDClD’15} dataset, and test on the entire \textit{NDIris3D-LG4000} and \textit{NDIris3D-AD100} datasets;


\item {\it regular-irregular tests:} In order to understand how well the methods generalize for unknown brands / lens patterns, we conduct the ``regular-irregular'' experiment on both \textit{NDIris3D-LG4000} and \textit{NDIris3D-AD100} datasets. We try to make the lens patterns from the training set and the testing set completely different. More specifically, we split the dataset into a {\it regular pattern} set, which consists of images of real irises and irises wearing contact lenses of regular (a dot-like) pattern, and an {\it irregular pattern} set, which consists of images of real irises and irises wearing contact lenses of irregular (not a dot-like) pattern. The definition of regular and irregular pattern was explained in Section~\ref{sec:NDCLD15}. The regular and irregular sets are subject-disjoint and brand-disjoint. Then, we first train on the regular set and test on the irregular set, and repeat the experiments with the training and testing sets swapped.
\end{enumerate}

\subsection{Results}
\subsubsection{train old, test new} 

The results are shown in Table~\ref{table:wacv_new}. Although there exist similar samples from the same brands in the training and testing sets, the aforementioned change in lens pattern over time makes the PAD task difficult\del{, even when the test images are acquired by the same LG4000 sensor}. Furthermore, when the test images are acquired by the AD100 sensor, the problem becomes even harder, as it is now additionally a cross-sensor evaluation. Accuracy of DACNN and RegionalPAD methods experiences a drop when the sensor is switched from LG4000 to AD100. This suggests that deep learning-based methods and methods that focus on local patches tend to overfit to the training dataset, performing poorly on unseen samples and samples from another sensor. The other three baselines achieve better accuracy but still drop for cross-sensor evaluation. \add{Note that OSPAD-3D has high BPCER and low APCER for LG4000 but low BPCER and high APCER for AD100. The reason is the following: although OSPAD-3D is not learning-based, it uses the score distributions in the training set to set a score threshold for test time. Therefore, this difference between sensors results from the different score distributions between images captured by LG4000 and AD100.} In contrast, OSPAD-fusion outperforms all other methods by a large margin, obtaining an accuracy of over 90\% for both sensors. Also, OSPAD-fusion is able to choose the correct predictions made by OSPAD-3D and OSPAD-2D. Compared to the higher accuracy of the two methods, OSPAD-fusion improves by 6.26\% on LG4000, and by 13.32\% on AD100.

\begin{table*}
\begin{center}
\caption{Performance of all iris PAD methods on newly collected data -- \textit{NDIris3D}.}
\label{table:wacv_new} 
\begin{tabular}{ccccccc}
\toprule
\multirow{3}{*}{Methods} & \multicolumn{6}{c}{Performance}\\
 & \multicolumn{3}{c}{LG4000} & \multicolumn{3}{c}{AD100} \\
\cline{2-7}
 & Acc. (\%) & APCER (\%) & BPCER (\%) & Acc. (\%) & APCER (\%) & BPCER (\%) \\
\midrule
OSPAD-fusion (this paper) & 94.76 & 6.36 & 4.12 & 91.14 & 11.84 & 5.92 \\ 
OSPAD-3D & 84.75 & 3.89 & 26.71 & 80.43 & 34.33 & 4.97 \\ 
OSPAD-2D & 89.18 & 21.27 & 0.46 & 79.54 & 40.19 & 0.95 \\ 
DACNN & 77.79 & 29.84 & 6.76 & 59.19 & 72.31 & 8.97 \\ 
SIDPAD & 80.61 & 31.10 & 3.90 & 73.35 & 49.76 & 3.76 \\ 
RegionalPAD & 78.37 & 39.91 & 3.44 & 56.30 & 64.24 & 23.3 \\ 
\bottomrule
\end{tabular}
\end{center}
\end{table*}

\subsubsection{regular-irregular tests} 

Table~\ref{table:regular_irregular_LG} presents the accuracy of the methods when we train them on the {\it regular} set and test on the {\it irregular} set for LG4000 sensor, and Table~\ref{table:irregular_regular_LG} shows the results when the train and test sets are swapped. From the tables, we see that OSPAD-3D achieves the highest accuracy among all open source PAD methods. Its performance is stable regardless of the training distribution. This matches with our hypothesis that OSPAD-3D does not rely on the textural information and its training only incorporates setting a score threshold, which is applied directly on variances of normal maps when testing. All other open source methods either perform poorly in both scenarios, or only manage to generalize in one of them, failing in the other scenario. On average, the only deep-learning-based method, DACNN, performs the worst, indicating that the method overfits on the training data. However, OSPAD-fusion is still able to improve the performance of OSPAD-3D through fusing with OSPAD-2D, though by a smaller margin. OSPAD-fusion is not influenced by the poor accuracy of OSPAD-2D, picking up only the accurate predictions.\par 

Similarly for the AD100 sensor, Table~\ref{table:regular_irregular_AD} presents the accuracy of the PAD methods when we train them on the {\it regular} set and test on the {\it irregular} set, and Table~\ref{table:irregular_regular_AD} shows the results when the train and test sets are swapped. The trends are similar to what we observed for the LG4000 sensor, with a very slight accuracy drop when trained on the {\it regular} test. It's interesting that DACNN again presents the worst accuracy. This shows that there may still be some effort required to make deep learning-based more effective in cross-domain scenarios.



\begin{table}
\begin{center}
\caption{Performance of all iris PAD methods on the LG4000 {\it irregular} set when trained on the {\it regular} set}
\label{table:regular_irregular_LG} 
\begin{tabular}{cccc}
\toprule
\multirow{2}{*}{Methods} & \multicolumn{3}{c}{Performance}\\ \cline{2-4}
 & Acc. (\%) & APCER (\%) & BPCER (\%)\\ 
\midrule
OSPAD-fusion (this paper) & 87.88 & 7.07 & 16.97 \\ 
OSPAD-3D & $87.17$ & $8.73$ & $16.77$ \\ 
OSPAD-2D & $62.42$ & $0.42$ & $73.25$ \\ 
DACNN & $78.05$ & $43.66$ & $1.10$ \\ 
SIDPAD & $79.99$ & $38.05$ & $2.69$ \\ 
RegionalPAD & $68.08$ & $3.59$ & $61.44$ \\ 
\bottomrule
\end{tabular}
\end{center}
\end{table}

\begin{table}
\begin{center}
\caption{Same as in Table \ref{table:regular_irregular_LG}, except that the methods were trained on the {\it irregular} set and tested on the {\it regular} set.}
\label{table:irregular_regular_LG} 
\begin{tabular}{cccc}
\toprule
\multirow{2}{*}{Methods} & \multicolumn{3}{c}{Performance}\\ \cline{2-4}
 & Acc. (\%) & APCER (\%) & BPCER (\%)\\ 
\midrule
OSPAD-fusion (this paper) & $91.40$ & $8.33$ & $8.87$ \\ 
OSPAD-3D & $87.93$ & $15.37$ & $8.87$ \\ 
OSPAD-2D & $73.49$ & $0.03$ & $57.36$ \\ 
DACNN & $59.26$ & $77.73$ & $2.55$ \\ 
SIDPAD & $63.23$ & $70.44$ & $2.02$ \\ 
RegionalPAD & $62.63$ & $5.64$ & $68.10$ \\ 
\bottomrule
\end{tabular}
\end{center}
\end{table}

\begin{table}
\begin{center}
\caption{Same as in Table \ref{table:regular_irregular_LG} but for the AD100 sensor.}
\label{table:regular_irregular_AD} 
\begin{tabular}{cccc}
\toprule
\multirow{2}{*}{Methods} & \multicolumn{3}{c}{Performance}\\ \cline{2-4}
 & Acc. (\%) & APCER (\%) & BPCER (\%)\\ 
\midrule
OSPAD-fusion (this paper) & $85.58$ & $5.81$ & $22.80$ \\ 
OSPAD-3D & $85.79$ & $6.24$ & $22.58$ \\ 
OSPAD-2D & $60.87$ & $2.30$ & $76.99$ \\ 
DACNN & $60.76$ & $67.63$ & $11.61$ \\ 
SIDPAD & $66.12$ & $63.23$ & $5.33$ \\ 
RegionalPAD & $63.26$ & $6.38$ & $67.96$ \\ 
\bottomrule
\end{tabular}
\end{center}
\end{table}

\begin{table}
\begin{center}
\caption{Same as in Table \ref{table:irregular_regular_LG} but for the AD100 sensor.}
\label{table:irregular_regular_AD} 
\begin{tabular}{cccc}
\toprule
\multirow{2}{*}{Methods} & \multicolumn{3}{c}{Performance}\\ \cline{2-4}
 & Acc. (\%) & APCER (\%) & BPCER (\%)\\ 
\midrule
OSPAD-fusion (this paper) & $82.79$ & $23.72$ & $10.63$ \\ 
OSPAD-3D & $77.78$ & $10.51$ & $34.06$ \\ 
OSPAD-2D & $75.34$ & $0.27$ & $48.79$ \\ 
DACNN & $56.10$ & $78.71$ & $8.72$ \\ 
SIDPAD & $69.51$ & $57.01$ & $3.68$ \\ 
RegionalPAD & $63.48$ & $6.54$ & $66.17$ \\ 
\bottomrule
\end{tabular}
\end{center}
\end{table}
\section{Conclusions}
In this paper, we first investigated the weaknesses of OSPAD-3D and showed that although it generalizes well to most unseen lens patterns, its assumptions are violated when the contact lens is heavily opaque, making the shadows less pronounced. In another set of experiments, we show that other existing iris PAD methods fail to generalize well to even changes in lens patterns made by the same manufacturers over time. To mitigate these limitations in existing iris PAD methods, we proposed a novel open source iris PAD method that combines 2D and 3D information from only two images, through a set of simple but extremely powerful fusion rules. We gathered all (known to us) available open source iris PAD methods, by either downloading codes from online repositories or contacting the corresponding authors of the papers. Extensive experimental results from the \textit{NDCLD'15} dataset and our newly collected \textit{NDIris3D} dataset show that the proposed method is robust under various open-set testing scenarios and outperforms all other iris PAD methods.
To facilitate reproducibility and directly comparable results, the source code and dataset are made available to the research community.


%



\section*{Data and Source Codes}

Upon acceptance of this paper, instructions on how to download a copy of the {\it NDIris3D} dataset will be posted to \textsf{https://cvrl.nd.edu/projects/data}, and the source codes of the proposed method will be posted to \textsf{https://github.com/cvrl}.

\ifCLASSOPTIONcompsoc
  \section*{Acknowledgments}
\else
  \section*{Acknowledgment}
\fi

The authors would like to thank Diane Wright for her help with \textit{NDIris3D} data collection. Also, the authors would like to thank Diego Gragnaniello, Yang Hu and Joseph McGrath for making their iris PAD source codes available for evaluation.

\ifCLASSOPTIONcaptionsoff
  \newpage
\fi



%

\bibliographystyle{IEEEtran}
\bibliography{refs}

%

\begin{IEEEbiography}[{\includegraphics[height=1.25in,clip,keepaspectratio]{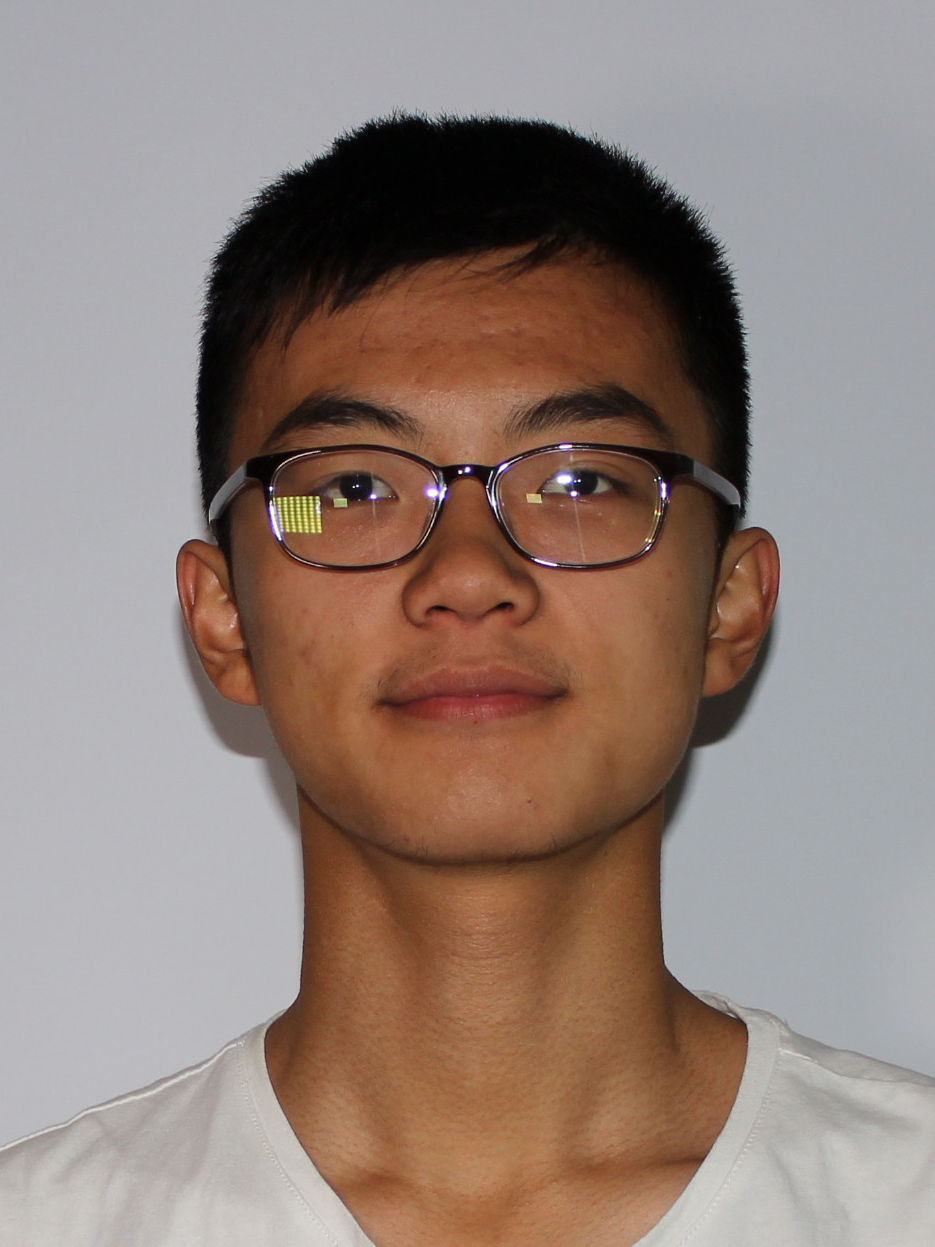}}]{Zhaoyuan Fang}(S'18) graduated from the University of Notre Dame with majors in Electrical Engineering and Mathematics in 2020. He is currently a research associate at the Notre Dame Computer Vision Research Lab. He will be a Master's student at Carnegie Mellon University. His current interests include iris presentation attack detections, multimodal learning and video/audio generation.
\end{IEEEbiography}

\begin{IEEEbiography}[{\includegraphics[height=1.25in,clip,keepaspectratio]{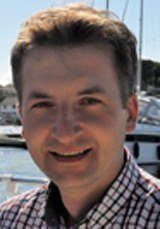}}]{Adam Czajka} (Ph.D. 2005, D.Sc. 2018) is an Assistant Professor in the Department of Computer Science and Engineering in the College of Engineering here at the University of Notre Dame. He is a Senior Member of the Institute of Electrical and Electronics Engineers, Inc. (IEEE) and VP for Finance of the IEEE Biometrics Council. Before coming to Notre Dame, Professor Czajka was the Chair of the Biometrics and Machine Learning Laboratory at the Institute of Control and Computation Engineering at WUT, the Head of the Postgraduate Studies on Security and Biometrics, the Vice Chair of the NASK Biometrics Laboratory, the Chair of the Polish Standardization Committee on Biometrics, an Assistant Professor in NASK – national research institute in Poland, and a member of the NASK Research Council.
\end{IEEEbiography}


\begin{IEEEbiography}[{\includegraphics[height=1.25in,clip,keepaspectratio]{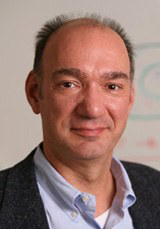}}]{Kevin W. Bowyer} is the Schubmehl-Prein Family Professor of Computer Science and Engineering at the University of Notre Dame, and also serves as Director of International Summer Engineering Programs for the Notre Dame College of Engineering. In 2019, Professor Bowyer was elected as a Fellow of the American Association for the Advancement of Science. Professor Bowyer is also a Fellow of the IEEE and of the IAPR, and received a Technical Achievement Award from the IEEE Computer Society, with the citation ``for pioneering contributions to the science and engineering of biometrics.'' Professor Bowyer currently serves as the Editor-in-Chief of the {\it IEEE Transactions on Biometrics, Behavior and Identity Science}, and previously served as Editor-in-Chief of the {\it IEEE Transactions on Pattern Analysis and Machine Intelligence}.
\end{IEEEbiography}




\end{document}